\documentclass[twocolumn,conference]{IEEEtran}
\usepackage[LGR,T1]{fontenc}
\usepackage{array}
\usepackage{booktabs}
\usepackage{enumitem}
\usepackage{calc}
\usepackage{textcomp}
\usepackage{multirow}
\usepackage{amsmath}
\usepackage{color}
\usepackage{balance}
\usepackage{graphicx,
psfrag,
epsfig,
epstopdf}
\usepackage[unicode=true,
 bookmarks=true,bookmarksnumbered=true,bookmarksopen=true,bookmarksopenlevel=1,
 breaklinks=false,pdfborder={0 0 0},pdfborderstyle={},backref=false,colorlinks=false]
 {hyperref}
\hypersetup{pdftitle={Your Title},
 pdfauthor={Your Name},
 pdfpagelayout=OneColumn, pdfnewwindow=true, pdfstartview=XYZ, plainpages=false}
\usepackage{breakurl}

\makeatletter

\DeclareRobustCommand{\greektext}{%
  \fontencoding{LGR}\selectfont\def\encodingdefault{LGR}}
\DeclareRobustCommand{\textgreek}[1]{\leavevmode{\greektext #1}}
\ProvideTextCommand{\~}{LGR}[1]{\char126#1}

\providecommand{\tabularnewline}{\\}

\makeatother

\begin{document}

\title{Bayesian Optimization with Machine Learning Algorithms Towards Anomaly Detection}

\author{\IEEEauthorblockN{MohammadNoor Injadat\IEEEauthorrefmark{1}, Fadi Salo\IEEEauthorrefmark{1},
Ali Bou Nassif\IEEEauthorrefmark{2}\IEEEauthorrefmark{1}, Aleksander Essex \IEEEauthorrefmark{1}, Abdallah Shami\IEEEauthorrefmark{1}}
\IEEEauthorblockA{\IEEEauthorrefmark{1}Department of Electrical and Computer Engineering,
University of Western Ontario, London, ON, Canada \\
Email: \{minjadat, fsalo, aessex, abdallah.shami\}@uwo.ca}
\IEEEauthorblockA{\IEEEauthorrefmark{2}Department of Computer Engineering,
University of Sharjah, Sharjah, UAE\\
Email: anassif@sharjah.ac.ae}}

\maketitle
\begin{abstract}
Network attacks have been very prevalent as their rate is growing
tremendously. Both organization and individuals are now concerned
about their confidentiality, integrity and availability of their critical
information which are often impacted by network attacks. To that end,
several previous machine learning-based intrusion detection methods
have been developed to secure network infrastructure from such attacks.
In this paper, an effective anomaly detection framework is proposed
utilizing Bayesian Optimization technique to tune the parameters of
Support Vector Machine with Gaussian Kernel (SVM-RBF), Random Forest
(RF), and k-Nearest Neighbor (k-NN) algorithms. The performance of
the considered algorithms is evaluated using the ISCX 2012 dataset.
Experimental results show the effectiveness of the proposed framework
in term of accuracy rate, precision, low-false alarm rate, and recall.
\end{abstract}


\IEEEpeerreviewmaketitle{}

\section{Introduction}
Computer networks and the Internet have become an essential component
of any organization in this high-tech world. Organizations heavily
depend on their networks to conduct their daily work. Moreover, individuals
are also dependent on the Internet as a means to communicate, conduct
business, and store their personal information \cite{bib1}. The topic
of Cyber-security has garnered significant attention as it greatly
impacts many entities including individuals, organizations, and governmental
agencies. Organizations have become more concerned with their network
security and are allocating more resources to protect it against potential
attacks or anomalous activities. Traditional network protection mechanisms
have been proposed such as adopting firewalls, authenticating users,
and integrating antivirus and malware programs as a first line of
defense \cite{bib2}. Nonetheless, these mechanisms have not been
as efficient in providing complete protection for the organizations'
networks, especially with contemporary attacks \cite{bib3}. 

Typical intrusion detection systems (IDSs) can be categorized into
two main types, namely signature-based detection systems (misused
detection) and anomaly-based detection systems \cite{bib4}. Signature-based
detection systems compare the observed data with pre-defined attack
patterns to detect intrusion. Such systems are effective for attacks
with well-known signatures and patterns. However, these systems miss
new attacks due to the ever-changing nature of intrusion attacks \cite{bib5}.
On the other hand, anomaly-based detection systems rely on the hypothesis
that abnormal behavior differs from normal behavior. Therefore, any
deviation from what is considered as normal is classified as anomalous
or intrusive. Such systems typically build models based on normal
patterns and hence are capable of detecting unknown behaviors or intrusions
\cite{bib6}. Although previous work on IDSs has shown promising improvement,
intrusion detection problem remains a prime concern, especially given
the high volume of network traffic data generated, the continuously
changing environments, the plethora of features collected as part
of training datasets (high dimensional datasets), and the need for
real-time intrusion detection \cite{bib7}. For instance, high dimensional
datasets can have irrelevant, redundant, or highly correlated features.
This can have a detrimental impact on the performance of IDSs as it
can slow the model training process. Additionally, choosing the most
suitable subset of features and optimizing the corresponding parameters
of the detection model can help improve its performance significantly
\cite{bib8}. 

In this paper, we propose an effective intrusion detection framework
based on optimized machine learning classifiers including Support
Vector Machine with Gaussian kernel (SVM-RBF), Random Forest (RF),
and k-Nearest Neighbors (k-NN) using Bayesian Optimization (BO). These
techniques have been selected based on the nature of the selected
dataset, i.e. SVM-RBF is selected because the data is not linearly
separable. Additional details about the utilized techniques are presented
in section III. This is done to provide a robust and accurate methodology
to detect anomalies. The considered methods are titled BO-SVM, BO-RF,
and BO-kNN respectively. The performance is evaluated and compared
by conducting different experiments with the ISCX 2012 dataset that
was collected from University of New Brunswick \cite{bib9}. As mentioned
in Wu and Banzhaf \cite{bib5}, a robust IDS should have a high detection
rate/recall and a low false alarm rate (FAR). Despite the fact that
most of intrusion detection methods have high detection rate (DR),
they suffer from higher FAR. Thus, this paper utilizes optimized machine
learning models to minimize the objective function that will maximize
the effectiveness of the considered methods. Totally, the feasibility
and efficiency of these optimized methods is compared using various
evaluation metrics such as accuracy (acc), precision, recall, and
FAR. Furthermore, the performance of the three optimized methods in
parameter setting are compared with the standard approaches. The main
contributions of this paper include the following: 
\begin{itemize}
	\item Investigate the performance of the optimized machine learning algorithms
	using Bayesian Optimization to detect anomalies. 
	\item Enhances the performance of the classification models through the
	identification of the optimal parameters towards objective-function
	minimization. 
	\item UNB ISCX 2012, a benchmark intrusion dataset is used for experimentation
	and validation purposes through the visualization of the optimization
	process of the objective function of the considered machine learning
	models to select the best approach that identifies anomalous network
	traffic. To the best of our knowledge, no previous related work has
	adopted Bayesian Optimization on the utilized dataset towards anomaly
	detection.
\end{itemize}
The remainder of this paper is organized as follows. Section II presents
the related work. Section III gives a brief overview of SVM, RF, and
k-NN algorithms along with the utilized optimization method. Section
IV discusses the research methodology and the experimental results.
Finally, Section V concludes the paper and provides future research
directions.

\section{Related Work}

The intrusion detection problem has been addressed as a classification
problem by researchers. Different data mining-based methodologies
have been posited to tackle this problem including, SVM \cite{bib10},
Decision Trees \cite{bib11}, k-NN \cite{bib12}, and Naive Bayes
\cite{bib13} classifiers as shown in the short review presented in
Tsai et al. \cite{bib1}. Later, noteworthy research have been implemented
and acquired promising results through proposing novel approaches
based data mining techniques Wu and Banzhaf \cite{bib5}. Recently,
many research adopted optimization techniques to improve the performance
of their approach. For instance, a hybrid approach proposed by Chung
and Wahid \cite{bib14} including feature selection and classification
with simplified swarm optimization (SSO). The performance of SSO was
further improved by using weighted local search (WLS) to obtain better
solutions from the neighborhood \cite{bib14}. Their experimental
results yielded accuracy of 93.3\% in detecting intrusions. Similarly,
Kuang et al. \cite{bib15} proposed a hybrid method incorporating
genetic algorithm (GA) and multi-layered SVM with kernel principal
component analysis (KPCA) to enhance the performance of the proposed
methodology. Another technique introduced by Zhang et al. \cite{bib17}
combining misuse and anomaly detection using RF. A novel algorithm
applied catfish effect named, Catfish-BPSO, had been used to select
features and enhance the model performance \cite{bib18}. Authors
used leave-one-out cross-validation (LOOCV) with k-NN for fitness
evaluation.

\section{THEORETIC ASPECTS OF THE TECHNIQUES}

\subsection{A. Support Vector Machines (SVM)}

SVM algorithm is a supervised machine learning classification technique
that identifies the class positive and negative sample by determining
the maximum separation hyperplane between the two classes \cite{bib19}.
Depending on the nature of the dataset, different kernels can be used
as part of the SVM technique since the kernel determines the shape
of the separating hyperplane. For example, a linear kernel can be
used in cases where the data is linearly separable by providing a
linear equation to represent the hyperplane. However, other kernels
are needed in cases where the data is not linearly separable. One
such kernel is the Gaussian Kernel . This kernel maps the data points
from their original input space into a high-dimensional feature space.
The output of the SVM with Gaussian kernel (also known as SVM-RBF)
is \cite{bib20}:
\begin{equation}
f(x)=w^{T}\Phi(x)+b
\end{equation}
where $\Phi(x)$ represents the used kernel. The goal is to determine
the weight vector $w^{T}$ and intercept $b$ that minimizes the following
objective function:
\small
\begin{equation}
\min_{w,b}\frac{1}{2} w^{2}+ C\sum_{i=1}^{m}\left[y_{i}\times cost_{1}(f(x_{i}))+(1-y_{i})\times cost_{0}(f(x_{i}))\right]
\end{equation}
\normalsize
where $C$ is a regularization parameter that penalizes incorrectly
classified instances, $cost_{i}$is the squared error over the training
dataset.

\subsection{k-Nearest Neighbors (k-NN)}

k-NN is a simple classification algorithm that determines the class
of an instance based on the majority class of its k nearest neighboring
points. This is done by first evaluating the distance from the data
point to all other points within the training dataset. Different distance
measures can be used such as the Euclidean distance or Mahalanoblis
distance. After determining the distance, the k nearest points are
identified and a majority voting-based decision is made on the class
of the considered data point \cite{bib21}. 

\subsection{Random Forests (RF)}

RF classifier is an ensemble learning classifier that combines several
decision tree classifiers to predict the class \cite{bib25}. Each
tree is independently and randomly sampled with their results combined
using majority rule. The RF classifier sends any new incoming data
point to each of its trees and chooses the class that is classified
by the most trees. RF algorithm works as follows \cite{bib22}: 
\begin{enumerate}
	\item Choose  $T$ number of trees to grow.
	\item Choose $m$ number of variables used to split each node.$m\ll M$,
	where $M$ is the number of input variables.
	\item Grow trees; While growing each tree, do the following:
	\begin{itemize}
		\item Construct a sample of size $N$ from $N$ training cases with replacement
		and grow a tree from this new sample. 
		\item When growing a tree at each node, select $m$ variables at random
		from $M$ and use them to find the best split.
		\item Grow tree to maximum size without pruning.
	\end{itemize}
	\item To classify point $X$, collect votes from every tree in the forest
	and then use majority voting to decide on the class label.
\end{enumerate}

\subsection{Bayesian Optimization (BO)}

Bayesian optimization algorithm \cite{bib23} tries to minimize a
scalar objective function $f(x)$ for $x.$ Depending on whether the
function is deterministic or stochastic, the output will be different
for the same input $x$. The minimization process is comprised of
three main components: a Gaussian process model for the objective
function $f(x)$, a Bayesian update process that modifies the Gaussian
model after each new evaluation of the objective function, and an
acquisition function $a(x).$ This acquisition function is maximized
in order to identify the next evaluation point. The role of this function
is to measure the expected improvement in the objective function while
discarding values that would increase it \cite{bib23}. Hence, the
expected improvement (EI) is calculated as:

\begin{equation}
EI(x,Q)=E_{Q}\big[\max(0,\mu_{Q}(x_{best})-f(x))\big]
\end{equation}
where $x_{best}$ is the location of the lowest posterior mean and
$\mu_{Q}(x_{best})$ is the lowest value of the posterior mean. 

\section{EXPERIMENTAL SETUP AND RESULT DISCUSSION}

\subsection{Dataset Description}

In this paper, the Information Security Centre of Excellence (ISCX)
2012 dataset was used to perform the experiments and evaluate the
performance of the proposed approach to detect anomalies. The entire
dataset comprises nearly 1.5 million network traffic packets, with
20 features and covered seven days of network activity (i.e. normal
and intrusion). Additional information about the dataset are available
in \cite{bib9}. A random subset has been extracted from the original
dataset. The training data contains 30,814 normal traces and 15,375
attack traces while the testing data contains 13,154 normal traces
and 6,580 additional attack traces.

\subsection{Experimental setup and Data Pre-processing}

The proposed techniques were implemented using MATLAB 2018a. Experiments
were carried out in an Intel\textregistered{} Core\texttrademark{}
i7 processor @ 3.40\,GHz system with 16GB RAM running Windows 10 operating
system. The selected dataset was transformed from their original format
into a new dataset consisting of 14 features. We eliminated the payload
features which include the actual packet as most of their contents
were empty, while start time, and end time features have been replaced
by duration feature. In the data normalization stage, attributes were
scaled between the range {[}0,1{]} by using Min-Max method to eliminate
the bias of features with greater values, the mathematical computation
is as follows: 

\begin{equation}
x'=\frac{x-\min(x)}{\max(x)-\min(x)}
\end{equation}
As most of the classifiers do not accept categorical features \cite{bib24},
data mapping technique was used to transform the non-numeric values
of the features into numeric ones, named \textit{categorical} in
MATLAB.

\subsection{Prediction Performance Measures}

To evaluate and compare prediction models quantitatively, the following
measurements were utilized: 

\begin{equation}
Accuracy=\frac{TP+TN}{TP+TN+FP+FN}
\end{equation}

\begin{equation}
Precision=\frac{TP}{TP+FP}
\end{equation}
\begin{equation}
Recall=\frac{TP}{TP+FN}
\end{equation}
where $TP$ is the true positive rate, $TN$ is the true negative
rate, $FP$ is the false positive rate, and $FN$ is the false negative
rate \cite{bib26}. 

\subsection{Results Discussion}

The aim of the work is to discover the optimized models\textquoteright{}
parameters of the utilized classifiers to classify the network intrusion
data with the selected parameters. The experimental scheme has been
done for each technique to reduce the cost function by tuning all
possible parameters to obtain the highest classification accuracy
and the minimum FAR. To that end, BO technique is used to determine
the optimal parameters for the considered machine learning models.
For instance, the optimal values of $C$ and $\gamma$ (for SVM),
the depth of trees and the adopted ensemble method (for RF), and the
value of $k$ and the distance measure method (for k-NN) are determined. 

For example, if we have a set of machine learning model parameters
$P*={P_{1},P_{2},\ldots,P_{n}}$ where $P_{i}$ is a parameter of
the parameters subset that needs tuning, then BO tries to minimize
the following cost function:

\begin{equation}
P*=\min\space J(P)
\end{equation}
where $J(P)$ is the associated cost function.

To visualize the behavior of the BO technique combined with the machine
learning technique on the training dataset, Figures 1 and 2 depict
how BO tunes the parameters towards the global minimum value of the
SVM cost function with respect to $C$ and $\gamma$ as parameters
subset. According to the figures, a unique global minimum is obtained
for $C=433.32$ and $\gamma=1.0586$. This in turn leads to improving
the model's training accuracy as shown in Table 1 from 99.58\% without
optimization to 99.95\% after optimization. Additionally, the testing
accuracy increases from 99.59\% to 99.84\%. On the other side, the
FAR had promising results with a reduction of 0.01 and 0.007 in the
training and testing datasets respectively. Table 2 also shows more
details about the optimization processing time.

\begin{figure}[htbp]
\centering
\includegraphics[scale=0.5]{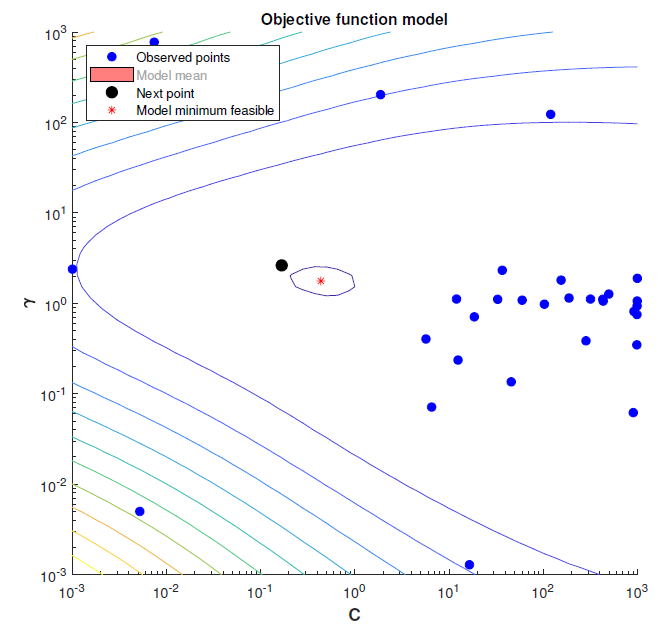}	
\caption{Optimized SVM Contour}
\label{ch6_fig1}
\end{figure}

\begin{table*}[tp]
	\caption{Performance results of the three classifiers}
	\centering{}%
	\begin{tabular}{lccccccccc}
		\toprule 
		& \multicolumn{4}{c}{Training} &  & \multicolumn{4}{c}{Testing}\tabularnewline
		\cmidrule{2-5} \cmidrule{7-10} 
		Classifier & Acc(\%) & Precision & Recall & FAR &  & Acc(\%) & Precision & Recall & FAR\tabularnewline
		\midrule
		SVM-RBF & 99.58 & 0.994 & 0.999 & 0.011 &  & 99.59 & 0.995 & 0.999 & 0.010\tabularnewline
		K-NN (k=5) & 99.59 & 0.9965 & 0.998 & 0.008 &  & 99.36 & 0.994 & 0.996 & 0.012\tabularnewline
		RF & 99.96 & 0.999 & 1.00 & 0.001 &  & 99.88 & 0.998 & 0.999 & 0.002\tabularnewline
		BO-SVM & 99.95 & 0.999 & 1.00 & 0.001 &  & 99.84 & 0.998 & 0.999 & 0.003\tabularnewline
		BO-k-NN & 99.98 & 0.999 & 1.00 & 0.001 &  & 99.93 & 0.999 & 0.999 & 0.001\tabularnewline
		BO-RF & 99.98 & 0.999 & 1.00 & 0.001 &  & 99.92 & 0.999 & 0.999 & 0.001\tabularnewline
		\bottomrule
	\end{tabular}
\end{table*}
\begin{table*}[tp]
	\caption{Optimization parametrs for each classifier}
	\centering{}%
	\begin{tabular}{|c|c|c|c|c|c|c|}
		\hline 
		\multirow{3}{*}{Best Parameters} & \multicolumn{2}{c|}{BO-SVM } & \multicolumn{2}{c|}{BO-k-NN} & \multicolumn{2}{c|}{BO- RF}\tabularnewline
		\cline{2-7} 
		& BoxConstraint (C) & 433.32 & NumNeighbors & 1 & Method & AdaBoost\tabularnewline
		\cline{2-7} 
		& KernelScale (\textgreek{g}) & 1.0586 & Distance & Mahalanobis & MaxNumSplits & 1004\tabularnewline
		\hline 
		Total function evaluations & \multicolumn{2}{c|}{30} & \multicolumn{2}{c|}{30} & \multicolumn{2}{c|}{30}\tabularnewline
		\hline 
		Total elapsed time in seconds & \multicolumn{2}{c|}{6175.78} & \multicolumn{2}{c|}{2272.50} & \multicolumn{2}{c|}{771.24}\tabularnewline
		\hline 
	\end{tabular}
\end{table*}

\begin{figure}[htbp]
	\centering
	\includegraphics[scale=0.5]{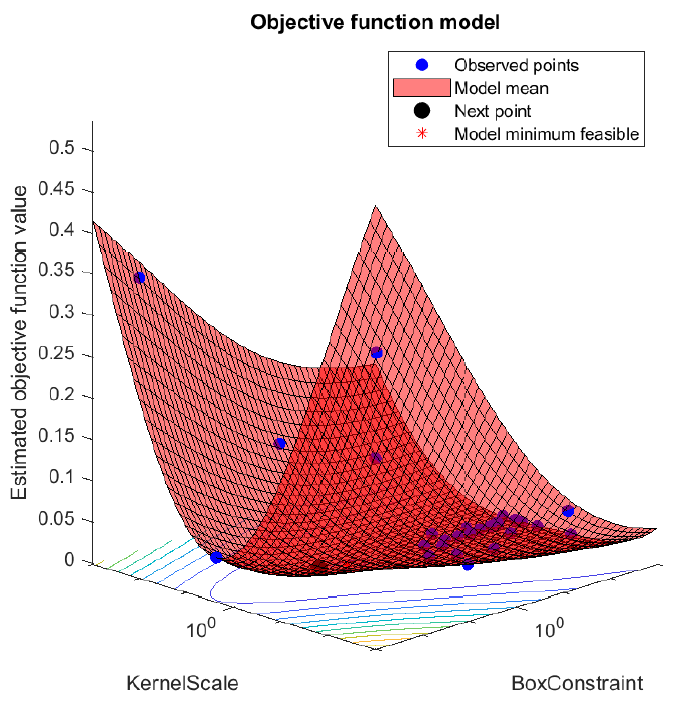}	
	\caption{Optimized SVM objective function model}
	\label{ch6_fig2}
\end{figure}

Similarly, Figures 3 and 4 and Table 2 show how the BO technique is
minimizing the cost function $J(P)$ for k-NN algorithm with respect
to the number of neighbors $k$ and the distance measuring method.
A unique global minimum is achieved for the values of $k=1$ and Mahalanobis
distance as the distance measuring method. According to Table 1, BO
was able to improve the BO performed 30 iterations to evaluate the
cost function in the aim to converge toward the optimal $J(P)$ of
each classifier.

\begin{figure}[htbp]
	\centering
	\includegraphics[scale=0.5]{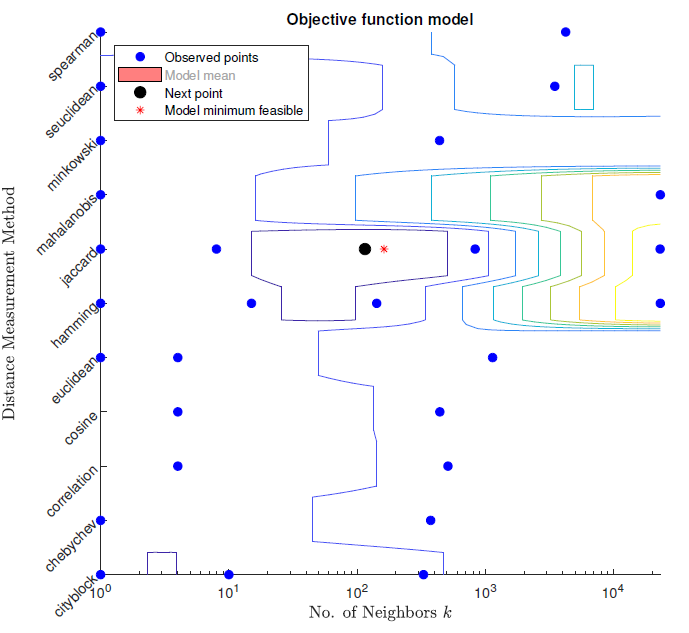}	
	\caption{Optimized k-NN Contour}
\end{figure}

\begin{figure}[htbp]
		\centering
		\includegraphics[scale=0.5]{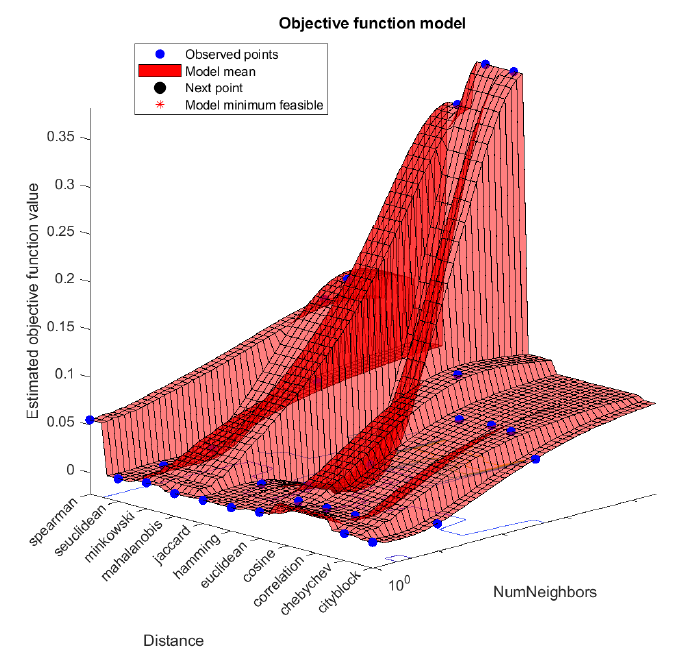}
	\caption{Optimized k-NN Objective Function Model }
\end{figure}

Figures 5, 6, and 7 visualize the change in the objective function
value vs the number of function evaluations for BO-SVM, BO-RF, and
BO-kNN respectively. It can be observed that the objective function
reaches its global minimum within 30 iterations at most. This reiterates
the efficiency of the BO technique in optimizing the considered algorithms.

\begin{figure}[htbp]
	\centering
	\includegraphics[scale=0.5]{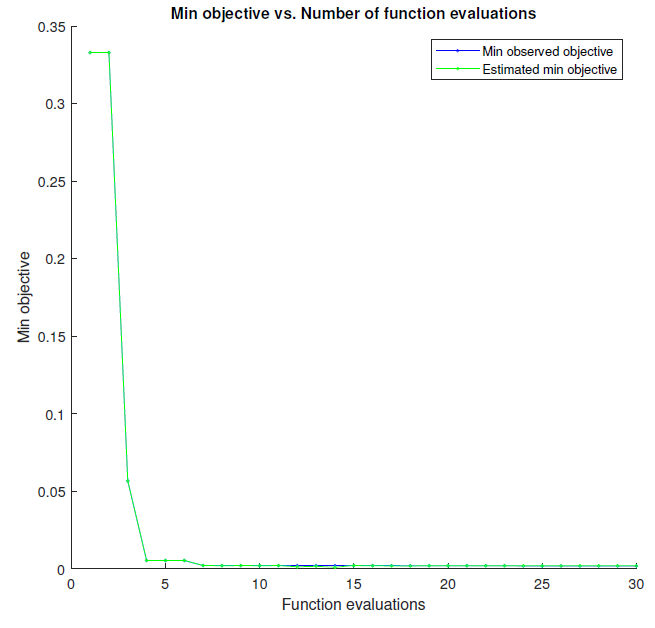}
	\caption{BO-SVM Objective Function vs Number of Function Evaluations}
\end{figure}

\begin{figure}[htbp]
	\centering
	\includegraphics[scale=0.38]{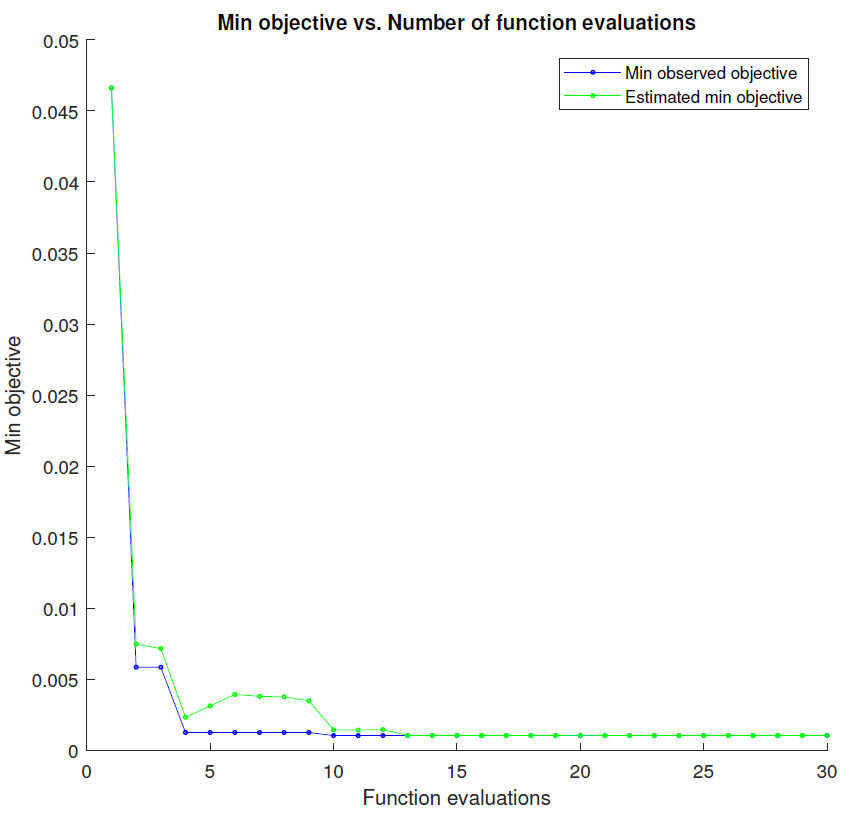}
	\caption{BO-kNN Objective Function vs Number of Function Evaluations}
\end{figure}

\begin{figure}[htbp]
		\centering
		\includegraphics[scale=0.38]{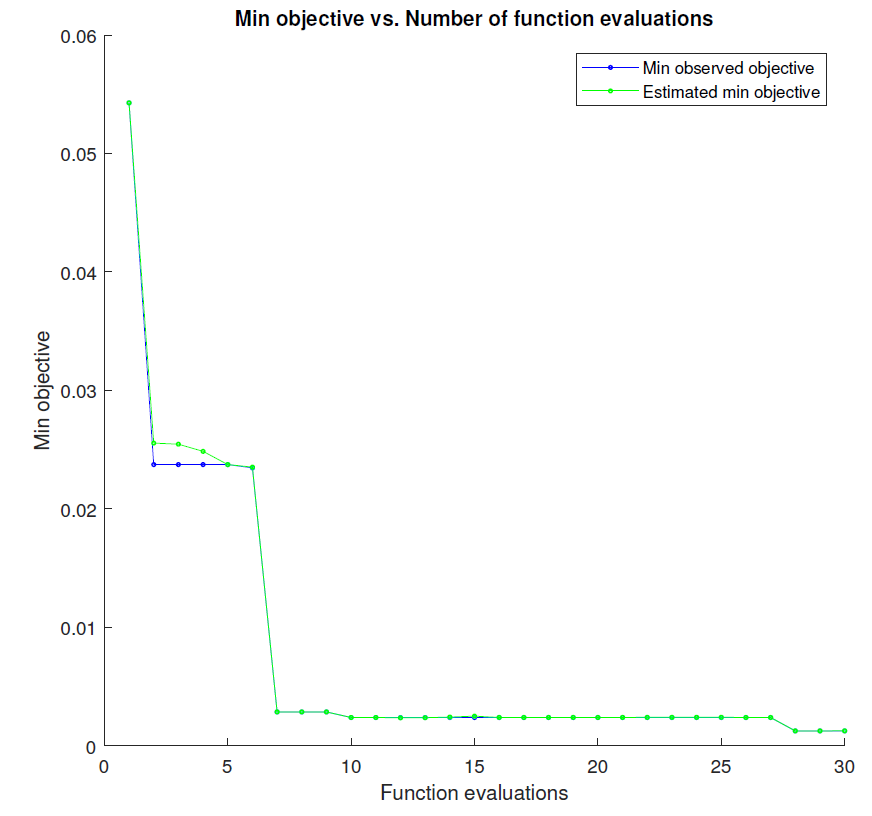}
	\caption{BO-RF Objective Function vs Number of Function Evaluations}
\end{figure}

By applying BO-RF, a unique global minimum is achieved with 1004 tree
splits (Tree Depth) and AdaBoost as a tree method. The BO improves
the training accuracy from 99.97\% to 99.98\% while the testing accuracy
improves from 99.88\% to 99.92\%. The FAR remains steady in the training
dataset and is reduced by 0.001 in the testing dataset. Furthermore,
Table 2 indicate that the BO find that AdaBoost is the best ensemble
method to build the tree.

It is also worth mentioning that Naïve Bayes classifier was utilized
at the initial stage of the experiment. However, due to the fact that
the dataset\textquoteright s features are not fully independent, the
classifier shows a low accuracy of 87.23\% and 87.65\% on the training
and testing datasets respectively. Hence, the Naïve Bayes classifier
was excluded from the experiment. 

Based on the previous publications, our results outperform the results
of previous experiments conducted using ISCX 2012 such as the results
shown in \cite{bib27} with their model acheiving about 95\% as overall
accuracy using their proposed technique. Additionally, \cite{bib28}
reported the highest accuracy of 99.8\% and 99.0\% for the training
and testing phases respectively.

\section{Conclusions}

In this paper, we utilized a Bayesian optimization method to enhance
the performance of anomaly detection methodology based on three conventional
classifiers; Support Vector Machine with Gaussian kernel (SVM-RBF),
Random Forest (RF), and k-Nearest Neighbor (k-NN). The BO optimization
method has been applied to set the parameters of these classifiers
by finding the global minimum of the corresponding objective function.
In order to have an efficient machine learning-based anomaly detection
system with high accuracy rate and a low false positive rate, BO was
able to improve the utilized classifiers. The experimental results
show not only is the proposed optimization method more accurate in
detecting intrusions, but also it can find the global minimum of the
objective function which leads to better classification results. Overall,
k-NN with Bayesian optimization has achieved the optimum performance
on ISCX 2012 dataset in terms of accuracy, precision, recall, and
false alarm rate. In order to further improve the performance of the
proposed approach, we plan to involve feature selection and parameter
setting applied simultaneously in the optimization method. Moreover,
the results of the proposed approach will be further improved by combining
both supervised and unsupervised machine learning techniques to detect
novel attacks with additional datasets such as the new release of
the ISCX dataset.
\balance
\bibliographystyle{IEEEtran}
\bibliography{bib1}

\end{document}